\documentclass[a4paper]{article}

\usepackage{INTERSPEECH2021}
\usepackage{pifont}
\usepackage{placeins}
\usepackage{cite}           
\usepackage[breaklinks=true,bookmarks=false,colorlinks=true]{hyperref}

\title{Spoken ObjectNet: A Bias-Controlled Spoken Caption Dataset}
\name{Ian Palmer, Andrew Rouditchenko\textsuperscript{\textdagger}\thanks{\textsuperscript{\textdagger} Corresponding author.}, Andrei Barbu, Boris Katz, James Glass\textsuperscript{\textdagger}}
\address{MIT CSAIL, USA}

\email{\{iapalm,roudi,abarbu,boris,glass\}@mit.edu}

\begin{document}

\maketitle
\begin{abstract}
Visually-grounded spoken language datasets can enable models to learn cross-modal correspondences with very weak supervision.  However, modern audio-visual datasets contain biases that undermine the real-world performance of models trained on that data. 
We introduce Spoken ObjectNet, which is designed to remove some of these biases and provide a way to better evaluate how effectively models will perform in real-world scenarios.  This dataset expands upon ObjectNet, which is a bias-controlled image dataset that features similar image classes to those present in ImageNet.

We detail our data collection pipeline, which features several methods to improve caption quality, including automated language model checks.  
Lastly, we show baseline results on image retrieval and audio retrieval tasks.  
These results show that models trained on other datasets and then evaluated on Spoken ObjectNet tend to perform poorly due to biases in other datasets that the models have learned.  We also show evidence that the performance decrease is due to the dataset controls, and not the transfer setting.

\end{abstract}

\section{Introduction}

Prior work in unsupervised spoken language learning has shown that neural models can learn meaningful audio-visual correspondences from visually grounded speech~\cite{harwath2016unsupervised,harwath2017learning,harwath2018jointly}.  
This mode of learning is inspired by humans in early childhood, who learn to use speech to describe the world before learning any written language. In practice, this could allow audio-visual models to learn from vast corpora of unlabeled images and videos.

However, many datasets in existence today, including audio-visual datasets, contain intrinsic biases that the models trained on those datasets then learn, which in turn degrades their performance on real-world data.
For example, most images and videos uploaded to the Internet are nicely lit, well-framed, and contain objects in their usual settings.  
In turn, image captioning models are biased towards describing people on beaches as happy and image classification models don't recognize wolves outside of a snowy backdrop \cite{objectrecognition}.

ObjectNet, a large-scale bias-controlled object classification dataset, addressed these problems by collecting a corpus of entirely new images instead of relying on those already uploaded to the Internet in some form \cite{objectnet}.  
Workers were asked to position a variety of household objects in a certain way against a specified background.  
The viewpoint of the camera was also controlled.  
In this way, ObjectNet has systematic controls in place for some of the biases that most other datasets exhibit.

In this work, we introduce Spoken ObjectNet (SON), a large-scale corpus of spoken image descriptions based on the ObjectNet dataset.
Our dataset addresses some of the biases present in existing audio-visual datasets.
We introduce our data collection pipeline, which includes a novel language modeling step that increases the general quality and acceptance rate of worker submissions.
Lastly, we conduct retrieval experiments to demonstrate that audio-visual models struggle to transfer to this bias-controlled domain, and the decreased performance is due to the controls and not just the transfer setting.
We will release the dataset publicly at \href{http://groups.csail.mit.edu/sls/downloads/son/index.cgi}{http://groups.csail.mit.edu/sls/downloads/son/index.cgi}, and our code is available at \href{https://github.com/iapalm/Spoken-ObjectNet}{https://github.com/iapalm/Spoken-ObjectNet}.

\section{Related Work}

\begin{figure*}[ht]
    \centering
    \includegraphics[width=\linewidth]{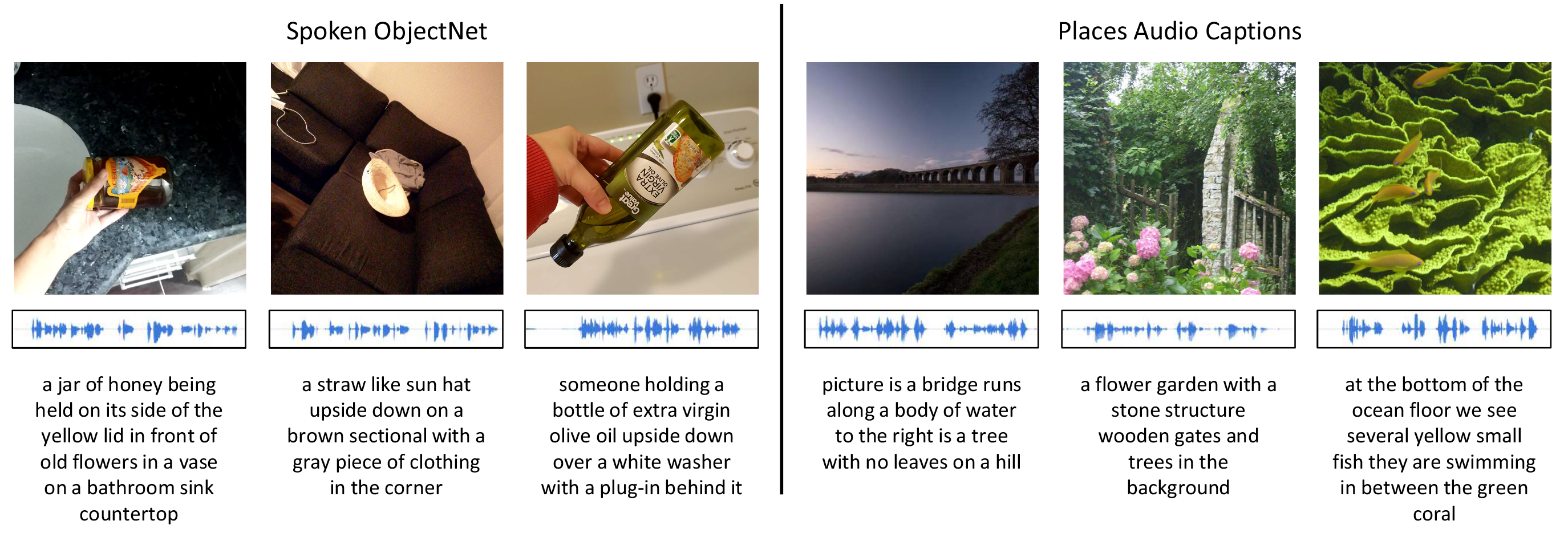}
    \caption{Samples of images, spoken captions, and ASR transcripts from Spoken ObjectNet and Places Audio~\cite{harwath2018jointly}.}
    \label{fig:samples}
\end{figure*}

\subsection{Spoken Caption Datasets}
Spoken captions of images were originally collected to build models that learn to recognize words from semantic-level supervision without any form of automatic speech recognition~\cite{harwath2015deep}.
Many spoken caption datasets have since been collected, and we show a comparison between them in Table~\ref{tab:compare}.
We focus on human-recorded captions in English, although spoken captions have been collected in Hindi~\cite{harwath2018vision} and Japanese~\cite{havard2019models,ohishi2020trilingual}.
The datasets mainly vary in modalities (images or videos) and whether the speech is spontaneous or read from text captions.

The Places Audio Captions dataset~\cite{harwath2018jointly} was the first large-scale dataset of spontaneous captions, and it contains over 400k spoken captions based on the Places 205 image dataset \cite{zhou2014learning}.  
The captions were collected via Amazon Mechanical Turk, with an average caption length of 20 words and audio duration of 10 seconds.  
Spoken ObjectNet shares a similar data collection framework and is approximately equal in average sequence length and duration.  
However, Spoken ObjectNet is smaller in scale than Places Audio Captions and also features the controls for bias within the ObjectNet dataset.  
Spoken ObjectNet, then, can function as a test set for models trained on Places Audio Captions.
Figure~\ref{fig:samples} shows samples from both datasets, demonstrating the stark differences in images and captions.

Localized Narratives~\cite{pont2020connecting} is also a recently collected large-scale spoken caption dataset with spontaneous speech.
Several other datasets exist, but they are based on previous text captions and therefore contain fewer words on average per caption.
The Flickr Audio Caption dataset was one of the first datasets and contains 40k read spoken captions based on the Flickr8k dataset and captions \cite{harwath2015deep, flickr8k}.
Spoken captions have been collected for the Microsoft Common Objects in Context (MS-COCO) dataset as well, which features over one million text captions for a variety of images~\cite{coco}. 
The SpokenCOCO~\cite{hsu2020textfree} dataset contains human-recorded captions, while other datasets contain synthetic captions generated using text-to-speech systems~\cite{havard2017speech,chrupala2017representations} which are less natural.
Synthetic spoken captions have also been collected in other contexts~\cite{ilharco2019large,Suris_2019_CVPR}.

Researchers have recently collected spoken captions for videos.
QuerYD~\cite{oncescu2021queryd} features audio descriptions for video segments annotated via YouDescribe, a volunteer-based community that adds narration to existing YouTube videos.
Spoken Moments~\cite{monfort2021spoken} contains spoken captions of 3s action video clips from Moments~\cite{monfort2019moments}.

Lastly, several video datasets exist without spoken captions but can still be used to learn audio-visual representations.
The How2~\cite{sanabria18how2} and HowTo100M~\cite{miech2019howto100m} instructional video datasets naturally contain spoken descriptions of visual events.
AudioSet~\cite{audioset} and VGG-Sound~\cite{chen2020vggsound} are both video datasets useful for audio-visual recognition tasks. 
Videos in VGG-Sound were selected and filtered using CNNs to curate a dataset with several hundred object classes.
Similarly, we use automated language modeling checks in our data collection pipeline.

\begin{table}[t]
  \centering
    \caption{Comparison of human-recorded spoken caption datasets in English. Mod.=Modality; W./C.=Words per Caption; S.=Spontaneous; C.=Control for Bias.}
    \vspace{-0.3cm}
    \resizebox{0.95\columnwidth}{!}{
    \begin{tabular}{@{}cccccc@{}}
      \toprule
      Dataset & Mod. & Samples & W./C. & S. & C.\\
      \cmidrule(r){1-1}\cmidrule(lr){2-2}\cmidrule(lr){3-3}\cmidrule(lr){4-4}\cmidrule(lr){5-5}\cmidrule(l){6-6}
      QuerYD \cite{oncescu2021queryd} & Video & 31,441 & 19.9 & \checkmark & \ding{55}\\
      Spoken Moments \cite{monfort2021spoken} & Video & 515,912 & 18 & \checkmark & \ding{55}\\
      Flickr Audio \cite{harwath2015deep} & Image &  40,000& 10.9 & \ding{55} & \ding{55}\\
      SpokenCOCO \cite{hsu2020textfree} & Image &  605,495 & 10 & \ding{55} & \ding{55}\\
      Loc. Narratives \cite{pont2020connecting} & Image & 873,107 & 36.5 & \checkmark & \ding{55}\\
      Places Audio \cite{harwath2018jointly} & Image &  402,385 & 21.9 & \checkmark & \ding{55}\\
      \textbf{Spoken ObjectNet} & Image & 50,273 & 20.5 & \checkmark & \checkmark\\
      \bottomrule
    \end{tabular}}
  \label{tab:compare}
\end{table}

\subsection{ObjectNet Dataset}
The ObjectNet dataset is an object detection test set collected in a way which explicitly controls for object viewpoints, rotations, and backgrounds.  
Removing these priors from the images results in significant performance drops (approximately 45\% for most models) versus performance on the ImageNet test set.  
The purpose of ObjectNet is to enable models that are more robust to real-world scenarios where objects may be in unusual contexts; similarly, the purpose of Spoken ObjectNet is to provide a test set for audio-visual models to measure how robustly they generalize to real-world situations.

\subsection{Audio-Visual Models}
Models for learning audio-visual correspondences typically learn an embedding space where true visual inputs and spoken caption pairs are close, while non-matching pairs are far apart.
In this work, we consider the ResDAVEnet~\cite{harwath2020jointly} architecture which combines CNN-based audio and image models.
The ResDAVEnet-VQ~\cite{Harwath2020Learning} architecture adds configurable vector quantization layers to the audio model.
Several other models that learn audio-visual correspondences from both images and videos have been presented in recent work~\cite{merkx2019language,boggust2019grounding,rouditchenko2020avlnet,sanabria2021talk,wang2021align}.

\section{Spoken ObjectNet Dataset Collection}

To collect samples for this dataset, we extended the approach used to collect the Places Audio Caption dataset~\cite{harwath2018jointly}.  
We released an Amazon Mechanical Turk (AMT) Human Intelligence Task (HIT), which allowed workers to submit captions for four images in the ObjectNet dataset at a time.  
Workers were compensated \$0.20 for four recordings that passed our validation steps.  
Workers were prohibited from submitting more than 3,000 HITs to prevent speaker bias from impacting the dataset.

During data collection, workers were given an image and asked to record themselves as if they were describing the image to someone who could not see it.  
Workers were told they could describe shapes, objects, locations, colors, and anything else of interest as they saw fit.  
After each recording was completed, we ran several validation steps on the recorded audio to ensure that it met our requirements.  
If a worker failed a validation step, they were immediately asked to redo the recording.  
We found that providing this feedback in real time, as opposed to rejecting the HIT outright hours or days later, increased the rate at which we could collect high-quality samples and improved the experience for workers.  
After four recordings were completed, workers could submit the assignment and proceed to the next HIT.

\subsection{Validation}
\label{sec:validation}

Each recording had to pass three checks in order for the worker to proceed to the next image.  
First, the recording had to be at least 1s in duration.  
This prevented workers from simply clicking through the screens as fast as possible in order to complete the task.  
The recording was also run through the Google Speech Recognition API to generate an ASR transcript of the recording.  
We required that each recording have at least four words in the transcript to be accepted.  
This prevented workers from recording silence or other non-speech sounds.

Lastly, we introduced a new step in which the ASR transcript was fed into a BERT model with a language modeling head.  
We used this model to produce a numerical score to approximate how well-formed the ASR text was.  
The model was a \verb|BertForMaskedLM| model from the Python Huggingface library \cite{wolf2020huggingfaces}, and our score is based on the cross entropy loss between the model's predictions based on the masked input tokens and the ground truth tokens.  
Any transcript that scores above a certain threshold (where higher scores are predicted to be less grammatical) failed the validation step.  
Given the unusual contexts of the objects and the potential for ASR errors, a low cutoff score could frustrate workers who were attempting to complete the task properly, so we used existing collected samples to measure a cutoff score that would prevent blatantly non-grammatical captions from passing.
Overall, this approach increased average caption quality, increased our HIT acceptance rate, and reduced the amount of manual validation that was required.

\subsection{Speaker Information}
\label{sec:speaker_id}
In practice, audio-visual models may learn information about the speaker's microphone instead of the content of the speech signal. 
If a worker annotated examples of primarily one class (e.g. all images from the \verb|measuring_cup| class), the model could exploit that correlation during training to make predictions without ever learning from the spoken words.  
To combat this, speakers were presented with images in a random order, so models cannot exploit speaker information to predict class identity.
The train, validation, and test splits were also constructed such that there is no speaker overlap between any of the sets. 

\subsection{Finalizing Splits}
In total, we collected over 70,000 samples.  
One sample per image in ObjectNet was selected to form the Spoken ObjectNet-50k dataset, with a total of 50,273 samples.  
48,273 are included in the training set, and 1,000 are included in both the validation and test sets.
We compare the vocabulary of the dataset with Places Audio~\cite{harwath2018jointly} in Table 2, where we find that our captions contain a similar part-of-speech distribution to a split of Places with 50k captions.

\begin{table}[t]
  \centering
    \caption{Vocabulary comparison of Spoken ObjectNet (SON-50k) and Places Audio (including our 50k split Places-50k).}
    \vspace{-0.3cm}
    \resizebox{0.95\columnwidth}{!}{
    \begin{tabular}{@{}ccccccc@{}}
      \toprule
      Dataset & Audio & Words & Nouns & Verbs & Adj. & Adv. \\
      \cmidrule(r){1-1}\cmidrule(lr){2-2}\cmidrule(lr){2-2}\cmidrule(lr){3-3}\cmidrule(lr){4-4}\cmidrule(lr){5-5}\cmidrule(lr){6-6}\cmidrule(l){7-7}
      \textbf{SON-50k} & 155h & 18,780 & 11,666 & 3,252 & 2,324 & 478  \\
      Places-50k & 115h & 20,140 & 11,212 & 4,332 & 2,963 & 620 \\
      Places-400k & 944h & 51,764 & 27,074 & 11,293 & 8,271 & 1,660 \\
      \bottomrule
    \end{tabular}}
  \label{tab:ch4compare}
\end{table}

\section{Retrieval Experiments}

\begin{figure}[t]
    \centering
    \includegraphics[width=\linewidth]{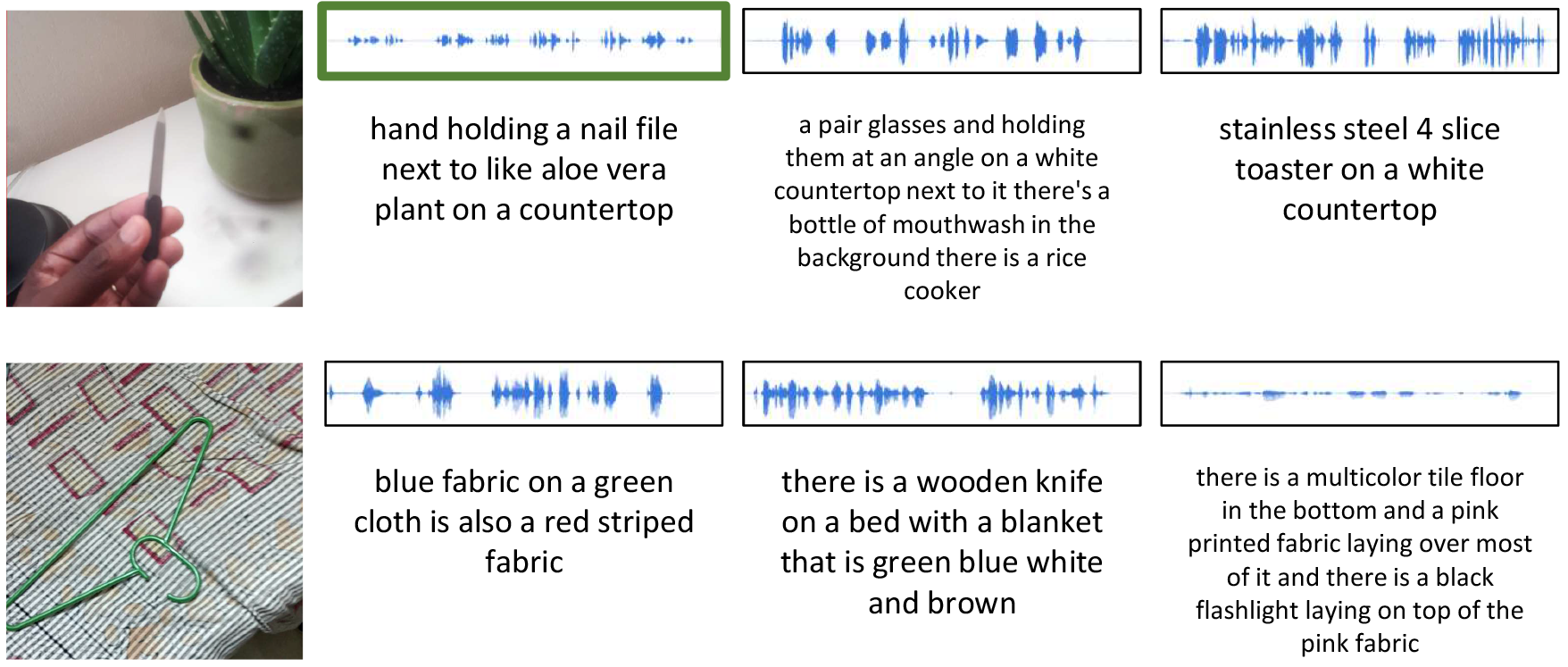}
    \caption{Top 3 retrieved audio captions for two sample images. The true caption for the image is boxed in green.}
    \label{fig:recalls}
\end{figure}

\begin{table*}[!htb]
  \centering
  \caption{Comparison of training on Spoken ObjectNet (SON) versus Places-50k.}
  \vspace{-0.3cm}
    \resizebox{0.9\linewidth}{!}{
    \begin{tabular}{@{}cccccccccccccc@{}}
      \toprule
      & & \multicolumn{6}{c}{Frozen image branch} & \multicolumn{6}{c}{Trainable image branch} \\
      \cmidrule(lr){3-8}\cmidrule(l){9-14}
      & & \multicolumn{2}{c}{I $\longrightarrow$ A} & \multicolumn{2}{c}{A $\longrightarrow$ I} & \multicolumn{2}{c}{Mean} & \multicolumn{2}{c}{I $\longrightarrow$ A} & \multicolumn{2}{c}{A $\longrightarrow$ I} & \multicolumn{2}{c}{Mean} \\
      \cmidrule(lr){3-4}\cmidrule(lr){5-6}\cmidrule(lr){7-8}\cmidrule(lr){9-10}\cmidrule(lr){11-12}\cmidrule(l){13-14}
      Dataset & Pretraining & R@1 & R@10 & R@1 & R@10 & R@1 & R@10 & R@1 & R@10 & R@1 & R@10 & R@1 & R@10 \\
      \cmidrule(r){1-1}\cmidrule(lr){2-2}\cmidrule(lr){3-3}\cmidrule(lr){4-4}\cmidrule(lr){5-5}\cmidrule(lr){6-6}\cmidrule(lr){7-7}\cmidrule(lr){8-8}\cmidrule(lr){9-9}\cmidrule(lr){10-10}\cmidrule(lr){11-11}\cmidrule(lr){12-12}\cmidrule(lr){13-13}\cmidrule(l){14-14}
      SON & ImageNet & 0.064 & 0.291 & 0.060 & 0.268 & 0.062 & 0.279 & 0.066 & 0.315 & 0.089 & 0.332 & 0.077 & 0.324 \\
      Places-50k & ImageNet & 0.093 & 0.364 & 0.079 & 0.360 & 0.086 & 0.362 & 0.067 & 0.306 & 0.081 & 0.335 & 0.074 & 0.321 \\
      SON & None & - & - & - & - & - & - & 0.017 & 0.123 & 0.016 & 0.132 & 0.017 & 0.128 \\
      Places-50k & None & - & - & - & - & - & - & 0.027 & 0.139 & 0.026 & 0.145 & 0.026 & 0.142 \\
      \bottomrule
    \end{tabular}}
  \label{tab:places50k}
\end{table*}

\begin{table}[t]
  \centering
  \caption{Transfer learning experiments from Places-400k to Spoken ObjectNet in the following settings: (1) No Fine-tuning (Zero-shot); (2) Fine-tuning (Frozen image branch); (3) Fine-tuning (Trainable image branch).}
  \vspace{-0.3cm}
  \resizebox{0.90\columnwidth}{!}{
    \begin{tabular}{@{}cccccccc@{}}
      \toprule
      & \multicolumn{2}{c}{I $\longrightarrow$ A} & \multicolumn{2}{c}{A $\longrightarrow$ I} & \multicolumn{2}{c}{Mean} \\
      \cmidrule(lr){2-3}\cmidrule(lr){4-5}\cmidrule(lr){6-7}
      Setting & R@1 & R@10 & R@1 & R@10 & R@1 & R@10 \\
      \cmidrule(r){1-1}\cmidrule(lr){2-2}\cmidrule(lr){2-2}\cmidrule(lr){3-3}\cmidrule(lr){4-4}\cmidrule(lr){5-5}\cmidrule(lr){6-6}\cmidrule(lr){7-7}
      (1) & 0.019 & 0.096 & 0.033 & 0.140 & 0.026 & 0.118 \\
      (2) & 0.040 & 0.216 & 0.048 & 0.213 & 0.044 & 0.214  \\
      (3) & 0.102 & 0.391 & 0.115 & 0.416 & 0.108 & 0.403 \\
      \bottomrule
    \end{tabular}}
  \label{tab:transfer}
\end{table}

\subsection{Experimental Setup}

Because Spoken ObjectNet is best understood as a test set relative to a dataset like Places Audio~\cite{harwath2018jointly}, our experiments are based on these two datasets.
We also created a split of Places Audio that is the same size as Spoken ObjectNet, with 48,273 training samples randomly selected from the original 400k and the same 1,000 sample validation set as the original Places Audio.
This split allows us to control for data quantity in our experiments, and we refer to the split as Places-50k and to the full training set as Places-400k.
Our primary experiments used the datasets' training sets to train models and the validation sets to evaluate models (as Places Audio did not have a predefined test set at the time these experiments were conducted).
We report results on the Spoken ObjectNet test set in Section~\ref{sec:test}.

We conducted both audio to image and image to audio retrieval experiments, where a model is tasked with retrieving the most similar images to a given audio caption, and vice versa.
We report two results, recall at 1 and recall at 10 (R@1 and R@10, respectively).  
For R@N, the model is successful if any of the top N recalled images are the correct match to the given audio caption (and vice versa for image to audio retrieval).

We tested frozen and trainable image branches, where freezing the image branch means prohibiting the parameters from being modified via backpropagation so that only the audio model and embedding layers are trained.
We define embedding layers as the light-weight layers (ie. linear or non-linear projections) after the backbone layers (ie. convolutional or attention layers).
The ObjectNet license prohibits training model parameters on images in the dataset, and the frozen setting complies with this restriction.  
We also compared image branches pretrained on ImageNet~\cite{imagenet} versus randomly initialized.

\subsection{Implementation Details}
All models were trained for 150 epochs with a batch size of 64 using the Adam optimizer~\cite{kingma2015adam}.
Every epoch we evaluated the model on the validation set, and the best performance out of the 150 epochs is reported. 
The learning rate depended on whether the image branch was frozen or not.  
For trainable image branch experiments, we use a learning rate of $2\cdot10^{-4}$ following~\cite{Harwath2020Learning}.
For frozen image branch experiments, a larger learning rate of $10^{-3}$ produced the best results.
Also, the learning rate exponentially decayed by a factor of 0.95 every 3 epochs.  
For all other hyperparameters and data processing, we followed~\cite{Harwath2020Learning}.

\subsection{Transfer from Places Audio to Spoken ObjectNet}

To understand how the bias controls in Spoken ObjectNet impact transfer performance from other spoken caption datasets, we ran transfer learning experiments with a model trained on Places Audio~\cite{Harwath2020Learning}.
We used the ResDAVEnet model trained on Places-400k which achieved a mean R@10 of 0.735 on the validation set.
There are two ways in which Spoken ObjectNet can be used as a test set: the first is for evaluating zero-shot performance (where the model undergoes no fine-tuning on Spoken ObjectNet), and the second is for evaluating performance after fine-tuning with a frozen image branch (where only the audio and embedding layers are fine-tuned).  
We also report the results of an experiment in which the entire image branch was made trainable and thus fine-tuned, strictly for comparison, as this setting will be prohibited due to ObjectNet's license.

The results are shown in Table~\ref{tab:transfer}.  
In the zero-shot setting, the model's mean R@10 performance decreases from 0.735 on Places to 0.118 on Spoken ObjectNet.
This shows that the model trained on Places can be directly applied to Spoken ObjectNet, but the performance is much lower.
Fine-tuning the model with a frozen image branch recovers some of the performance, up to a 0.214 mean R@10.  
When the image branch is made trainable, the performance increases to a mean R@10 of 0.403.  
These experiments show that the controls for viewpoint, rotation, and background make it difficult for the image model (trained on Places-400k) to meaningfully featurize the images in Spoken ObjectNet, as fine-tuning the embedding layers and audio model without fine-tuning the entire image model was not enough to recover the performance of the fully-trainable model.

\subsection{Comparison of Spoken ObjectNet and Places Audio}
Table \ref{tab:places50k} compares the relative difficulties of Places-50k and Spoken ObjectNet (SON), where the datasets are matched in size.  By running these experiments, we provide additional evidence that the difficulty of Spoken ObjectNet (and the performance drop shown in the transfer setting) is due to the controls for bias.  In the frozen image branch setting, the model trained on Spoken ObjectNet performs significantly worse than the model trained on Places-50k based on mean R@10.  
These results indicate that the ImageNet-pretrained image model is more effective for Places-50k than Spoken ObjectNet when it is kept frozen.

In the second half of Table \ref{tab:places50k}, we show the results of two pairs of experiments in which the image branch was trainable.  
While this setting will be prohibited due to ObjectNet's license, we show the results to give insight on the difficulty of Spoken ObjectNet versus Places.
In the first experiment, the image branch was pretrained on ImageNet.  In this experiment, the performance of the model trained on Spoken ObjectNet increases by approximately 20\% relative to its frozen counterpart.  However, the model trained on Places-50k with a trainable image branch actually decreases in performance compared to the frozen image branch model.  This decrease in performance is surprising, and as a result the mean R@10 scores of both models are roughly equivalent.  This is likely due to the class overlap between ImageNet and ObjectNet.  With a relatively small number of training samples, the model is able to learn enough about the viewpoint, rotation, and background controls applied to objects it already knows about to increase its performance.  On the other hand, when the parameters of the Places-50k image model are adjusted on a relatively small set of Places images it results in a featurizer that performs worse than the original frozen ImageNet-pretrained model.

In the second experiment, the image branch was still fully trainable, but not pretrained on ImageNet.
The model trained on Places-50k slightly outperforms the model trained on Spoken ObjectNet, but by a small margin.
This shows that without any prior training on any other datasets, and thus without leveraging biases learned from other datasets, Spoken ObjectNet and Places-50k are comparable in difficulty to learn from.

Finally, we show qualitative retrieval results in Figure~\ref{fig:recalls} for the model trained and evaluated on Spoken ObjectNet.  
In these examples, the model retrieves several relevant captions for each image, including the true caption for the first image.
However, the relevance of the other captions could be improved.

\begin{table}[t]
  \centering
  \caption{Experiments on the Spoken ObjectNet test in the following settings: (1) Zero-shot from Places-400k; (2) Fine-tuning (Frozen image branch) from Places-400k; (3) Training from scratch (Trainable image branch).}
  \vspace{-0.3cm}
  \resizebox{0.90\columnwidth}{!}{
    \begin{tabular}{@{}cccccccc@{}}
      \toprule
      & \multicolumn{2}{c}{I $\longrightarrow$ A} & \multicolumn{2}{c}{A $\longrightarrow$ I} & \multicolumn{2}{c}{Mean} \\
      \cmidrule(lr){2-3}\cmidrule(lr){4-5}\cmidrule(lr){6-7}
      Setting & R@1 & R@10 & R@1 & R@10 & R@1 & R@10 \\
      \cmidrule(r){1-1}\cmidrule(lr){2-2}\cmidrule(lr){2-2}\cmidrule(lr){3-3}\cmidrule(lr){4-4}\cmidrule(lr){5-5}\cmidrule(lr){6-6}\cmidrule(lr){7-7}
      (1) & 0.024 & 0.108 & 0.034 & 0.139 & 0.029 & 0.123 \\
      (2) & 0.034 & 0.201 & 0.031 & 0.196 & 0.033 & 0.199  \\
      (3) & 0.074 & 0.291 & 0.062 & 0.273 & 0.068 & 0.282 \\
      \bottomrule
    \end{tabular}}
  \label{tab:test}
\end{table}

\subsection{Spoken ObjectNet Test Set Experiments}
\label{sec:test}
Since Places Audio did not have a test set at the time of conducting the experiments, our main experiments compare Spoken ObjectNet and Places Audio using their validation sets.
However, for future work, we also provide a subset of the results evaluated on the Spoken ObjectNet test set.
The results were obtained by selecting the best performing model weights on the validation set and evaluating the models with those weights on the test set. 
The results are shown in Table~\ref{tab:test}.
We evaluate the model trained on Places-400k in the zero-shot setting and in the fine-tuned (frozen image branch) settings, which correspond to the results in the first two rows of Table~\ref{tab:transfer}.
We also evaluate the model trained from scratch on Spoken ObjectNet (frozen image branch), which corresponds to the results in the top left row of Table~\ref{tab:places50k}.
As a general observation, the results on the test set are similar to the results on the validation set.
We expect these results to serve as the baselines for future work.

\section{Conclusion}

We introduce Spoken ObjectNet as a bias-controlled spoken language dataset designed to function as a ``test set" for audio-visual models.
To use the dataset, we suggest training an audio-visual model on some other dataset first.  To evaluate the performance of the model in a bias-controlled setting, evaluate the model on the provided 1,000 sample test set.  To account for the different classes in ObjectNet and to therefore improve performance slightly, the model's embedding layers and audio model may be fine-tuned on the Spoken ObjectNet training set.  As with the original ObjectNet dataset, training model parameters on the images is prohibited.

Spoken ObjectNet exposes the performance gains that models gain from the priors baked into many modern datasets.
Our hope is that Spoken ObjectNet can provide inspiration for researchers to explore model architectures that are more robust to priors in data and therefore more likely to generalize to real-world scenarios.

\section{Acknowledgements}
We thank Rami Manna for helpful discussions. This research was sponsored by the United States Air Force Research Laboratory and the United States Air Force Artificial Intelligence Accelerator and was accomplished under Cooperative Agreement Number FA8750-19-2-1000. Andrei Barbu and Boris Katz were also supported by the Center for Brains, Minds and Machines, NSF STC award 1231216, and the Office of Naval Research under Award Number N00014-20-1-25. The views and conclusions contained in this document are those of the authors and should not be interpreted as representing the official policies, either expressed or implied, of the United States Air Force or the U.S. Government. The U.S. Government is authorized to reproduce and distribute reprints for Government purposes notwithstanding any copyright notation herein.

\bibliographystyle{IEEEtran}

\bibliography{template}

\end{document}